%% file: main.tex
\documentclass[conference]{IEEEtran}
\usepackage{times}

\usepackage[numbers]{natbib}
\usepackage{multicol}
\usepackage[bookmarks=true]{hyperref}
\usepackage{amsmath}
\usepackage{amssymb}
\usepackage{graphicx}
\usepackage{wrapfig}

\let\labelindent\relax
\usepackage{enumitem}

\usepackage{algorithm}
\usepackage[noend]{algpseudocode}
\usepackage{varwidth}

\usepackage{booktabs}

\begin{document}

\title{Improvisation through Physical Understanding: Using Novel Objects as Tools with Visual Foresight}

\author{\authorblockN{Annie Xie, Frederik Ebert, Sergey Levine, Chelsea Finn}
\authorblockA{UC Berkeley
}
}


%

\maketitle

\begin{abstract}
Machine learning techniques have enabled robots to learn narrow, yet complex tasks and also perform broad, yet simple skills with a wide variety of objects. However, learning a model that can both perform complex tasks and generalize to previously unseen objects and goals remains a significant challenge. We study this challenge in the context of ``improvisational'' tool use: a robot is presented with novel objects and a user-specified goal (e.g., sweep some clutter into the dustpan), and must figure out, using only raw image observations, how to accomplish the goal using the available objects as tools. We approach this problem by training a model with both a visual and physical understanding of multi-object interactions, and develop a sampling-based optimizer that can leverage these interactions to accomplish tasks. We do so by combining diverse demonstration data with self-supervised interaction data, aiming to leverage the interaction data to build generalizable models and the demonstration data to guide the model-based RL planner to solve complex tasks. Our experiments show that our approach can solve a variety of complex tool use tasks from raw pixel inputs, outperforming both imitation learning and self-supervised learning individually. Furthermore, we show that the robot can perceive and use novel objects as tools, including objects that are not conventional tools, while also choosing dynamically to use or not use tools depending on whether or not they are required. Videos of the results are available online\footnote{The project website is at \url{https://sites.google.com/view/gvf-tool}}.
\end{abstract}

\IEEEpeerreviewmaketitle

\input{introduction.tex}

\input{related_work.tex}

\input{overview.tex}

\input{method}

\input{experiments}

\section{Discussion} 
\label{sec:conclusion}

\noindent \textbf{Summary.}
We developed an approach to enable a robot to accomplish both \emph{diverse} and \emph{complex} tasks involving previously-unseen objects with access to only raw visual inputs. We studied the particular case of solving many different tasks that require manipulating objects as tools. Our approach learns from a combination of diverse human demonstration data, with many different goals, tools, and items, and autonomously-collected interaction data, with diverse objects. We show how we can use this data to train a model that can predict the visual outcome of actions that cause multi-object interaction, and use these predictions to figure out how to accomplish tasks by leveraging such object-object interactions.

\noindent \textbf{Limitations and Future Work.}
Our approach has a number of limitations that we hope to study in future work.
First, the tool-use tasks that we consider are diverse, but largely involve sweeping, wiping, and hooking interactions. 
In future work, we hope to also study tool use problems that involve cutting, skewering, and screwing interactions. In these cases, we expect that a more unconstrained action space may be important, where demonstrations may be of even greater importance to direct exploration within the larger state space. Second, our approach uses entirely visual observations, while a number of tool-use applications, such as using a screwdriver, demand force feedback. 
In principle, our approach abstracts away the form of the observation through learning. Indeed, prior work has shown that approaches like visual foresight can be integrated with tactile sensor inputs~\cite{tian2019manipulation}; but in practice, the introduction of tactile or force sensors would likely introduce additional challenges in evaluating predictions and collecting data safely.
Finally, while tool use provides an interesting test-bed for studying diverse, yet complex manipulation problems, we hope to study our approach in the context of other temporally-extended skills in future work.

\section*{Acknowledgments}

We gratefully thank Stephen Tian for creating many of the tools used in our experiments. This work was supported by the National Science Foundation through IIS-1651843, IIS-1700697, and IIS-1700696, the Office of Naval Research, Amazon, Google, and NVIDIA.


\bibliographystyle{plainnat}
\bibliography{references}

\input{appendix.tex}

\end{document}

%% file: introduction.tex
\section{Introduction}

An understanding of physical cause-and-effect relationships is a powerful means for enabling robots to achieve a wide variety of complex goals. 
This understanding becomes especially useful when performing complex multi-object manipulation tasks, such as those involved in tool use: if a robot could predict how one object might interact with another, it would be able to autonomously construct tool-use behaviors on the fly.
While fully-specified analytic and symbolic models of physics can allow fully observable systems to perform such tasks~\cite{toussaint2018differentiable}, acquiring such models is substantially more challenging when the environment can only be observed through image observations. Learning predictive models of low-level observations, such as camera image pixels, has a number of benefits. Such models can be learned from real world data and deployed in real world settings, as they do not require direct access to the state of the objects in the world. Models from pixels
further do not need knowledge of object shapes, surface friction, or other properties, and hence can use large datasets of experience and readily generalize to new objects. Indeed, model-based reinforcement learning with action-conditioned video prediction models, known as \emph{visual MPC, or visual foresight}, has enabled robots to perform short-horizon tasks involving a range of novel, previously-unseen objects~\cite{finn2017deep,ebert2018visual,ebert2017self}.

\begin{figure}[!t]
    \centering
    \includegraphics[width=\linewidth]{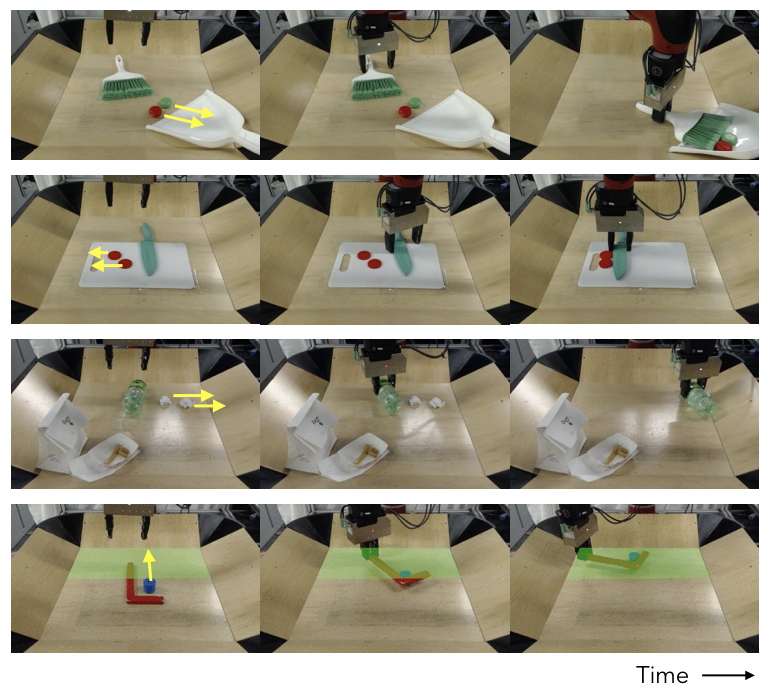}
    \vspace{-0.5cm}
    \caption{The robot learns a visual predictive model and uses it to manipulate new objects, that were never seen before, as tools to accomplish tasks specified by a person (as indicated by the yellow arrows). The robot can even utilize objects that are not conventional tools, such as water bottles. In the last example, when the robot is allowed to move only within the shaded green region, it uses the L-shaped hook to pull the blue cylinder towards itself.
    }
    \vspace{-0.3cm}
    \label{fig:teaser}
\end{figure}

In this paper, we investigate how models that predict low-level percepts, such as future camera images, can enable a robot to reason about multi-object interactions. In particular, we aim to study ``improvisational'' tool use, where the robot can use new objects, which might never have been seen before, to interact with other items to perform tasks that are not achievable with only basic single-object manipulation skills (e.g., grasping and pushing with the robot's own end-effector).  We focus on tool use in this work because it represents one of the most fundamental multi-object manipulation capabilities, and is a skill that is often associated with greater levels of intelligence in animals~\cite{osiurak2016tool}. 

Imitation learning approaches have enabled robots to learn to perform complex tasks~\cite{atkeson1997robot,argall2009survey}, including some types of tool use~\cite{rajeswaran2017learning},
while visual MPC has enabled robots to perform skills with many different novel objects~\cite{finn2017deep,ebert2018visual}. However, both have their limitations: policies learned through imitation are typically inflexible, as they are constrained to imitate the demonstrator, while work on visual MPC~\cite{finn2017deep,ebert2018visual} has so far been limited to simple, short-term skills. We combine ideas from imitation learning and from visual MPC to show that their combination can outperform each approach individually when applied to problems requiring tool use. In particular, we show that we can use demonstrations to solve complex tasks, while retaining the flexibility of visual planning.

The main contribution of this paper is a study of how direct prediction of low-level sensory observations, namely camera images, can enable a robot to carry out improvised multi-object interactions -- that is, determine how to use tools in its environment to perform tasks that require tool use. To this end, we combine ideas from imitation learning and prior work on visual MPC~\cite{finn2017deep}, incorporating imitation-driven models into both the data collection process and the sampling-based planning procedure. Our method uses video prediction to reason about potential robot actions, constructing plans to manipulate novel objects on the fly, in less than a second. Our experiments with a real-world Sawyer robot indicate that, by leveraging demonstrations and autonomously collected, self-supervised data, the robot can decide to use tools in situations where they are needed and use its arm directly, without the tool, in situations where tools are not needed.
Further, by reasoning about object interactions, the robot can find effective tool-use strategies even if it has never seen the tool before, and even in situations where no conventional tools are available. In comparisons, our approach exceeds the performance of both direct imitation and direct visual planning.

%% file: related_work.tex
\section{Related Work}

We discuss prior approaches to tool use along with robotic learning methods that use demonstrations or video prediction.

\noindent \textbf{Planning tool use with analytic models.}
Robotic manipulation involving tools has been studied in the task and motion planning (TAMP) literature~\cite{kuffner2005motion, latombe2012robot, halperin2000general, yamashita1998cooperative, gupta1998micro, mordatch2012discovery}.
\cite{toussaint2018differentiable, brown2012relational} propose to use logic programming together with known models to algorithmically discover tool-use. One challenge that limits the scalability of most logic-based systems and analytic model-based systems is that modeling errors quickly accumulate during execution, which often results in fragile system. Unlike this prior work, we study tool use in the real world with visual inputs, using learned dynamics models and sampling-based optimization.

\noindent \textbf{Direct learning methods for tool use.} Several works have decomposed tool use into multiple stages~\cite{brown2012relational, tikhanoff2013exploring}: tool selection, task-oriented grasping of the selected tool~\cite{li1988task,shimoga1996robot,bekiroglu2013probabilistic,antonova2018global,fang2018learning}, and using the grasped tool through planning~\cite{deepmpc} or policy learning~\cite{fang2018learning}. These methods constrain the scope of motions to trajectories that involve the tool, while our method is capable of finding plans with or without a tool based on the situation (see, e.g., Figure~\ref{fig:no_tool_use}). Other approaches have learned dynamics models to predict the outcome of applying actions to a tool~\cite{stoytchev2005behavior,deepmpc}. Unlike these approaches, which either use hand-designed perception pipelines or no visual perception at all, our method learns about object interactions directly from raw image pixels, avoiding restrictive assumptions that might hinder generalization.

\noindent \textbf{Learning from demonstrations.} Imitating expert demonstrations is a common approach for learning complex skills and can overcome the exploration challenges arising in long-horizon control problems~\cite{argall2009survey}. 
Prior work has leveraged demonstrations to accelerate \emph{model-free reinforcement learning} either in simulation or the real world, overcoming the well-known exploration problem~\cite{rajeswaran2017learning, nair2018overcoming, kober2011reinforcement, kormushev2010robot}. Unlike most of these works, our method does not fully rely on a policy to obtain actions.
Further, our method can leverage demonstrations of many different tasks and goals, by combining a stochastic policy learned from imitation with goal-directed model-based planning. 
Learning from demonstrations has also been used in combination with planning , where a planning cost function is inferred from data~\cite{2018arXiv181006544R, ye2017guided, aleotti2006grasp, lawitzky2012feedback} or where tool-use is learned from human demonstrations using pose-tracking~\cite{zhu2015understanding, lee2008association}. These approaches can be brittle since accurate pose-tracking of objects is often challenging in real-world scenarios.

\noindent \textbf{Video prediction-based planning}
Our work extends visual MPC~\cite{finn2017deep,ebert2018visual}, also referred to as visual foresight, which is a model-based reinforcement learning approach where a deep neural network model is trained to predict future visual observations. Such methods have been used in prior work for reaching~\cite{byravan2017se3}, pushing objects~\cite{finn2017deep,ebert2017self}, basic grasping and relocation~\cite{ebert2018robustness}, and manipulating clothes~\cite{ebert2018visual}. Our aim is to extend these methods to enable improvisational tool use.
Visual MPC, as well as other video prediction-based planners \cite{byravan2017se3, finn2017deep, paxton2018visual, watter2015embed}, generally do not succeed at such temporally extended tasks. To this end, we propose to incorporate demonstrations into the algorithm to enable  multi-stage tool-use capabilities, while still retaining the flexibility of goal-directed planning to accomplish varied user-specified goals.

%% file: overview.tex
\section{Capabilities for Improvisational Tool Use}

\begin{figure*}[t]
    \centering
    \includegraphics[width=0.8\linewidth]{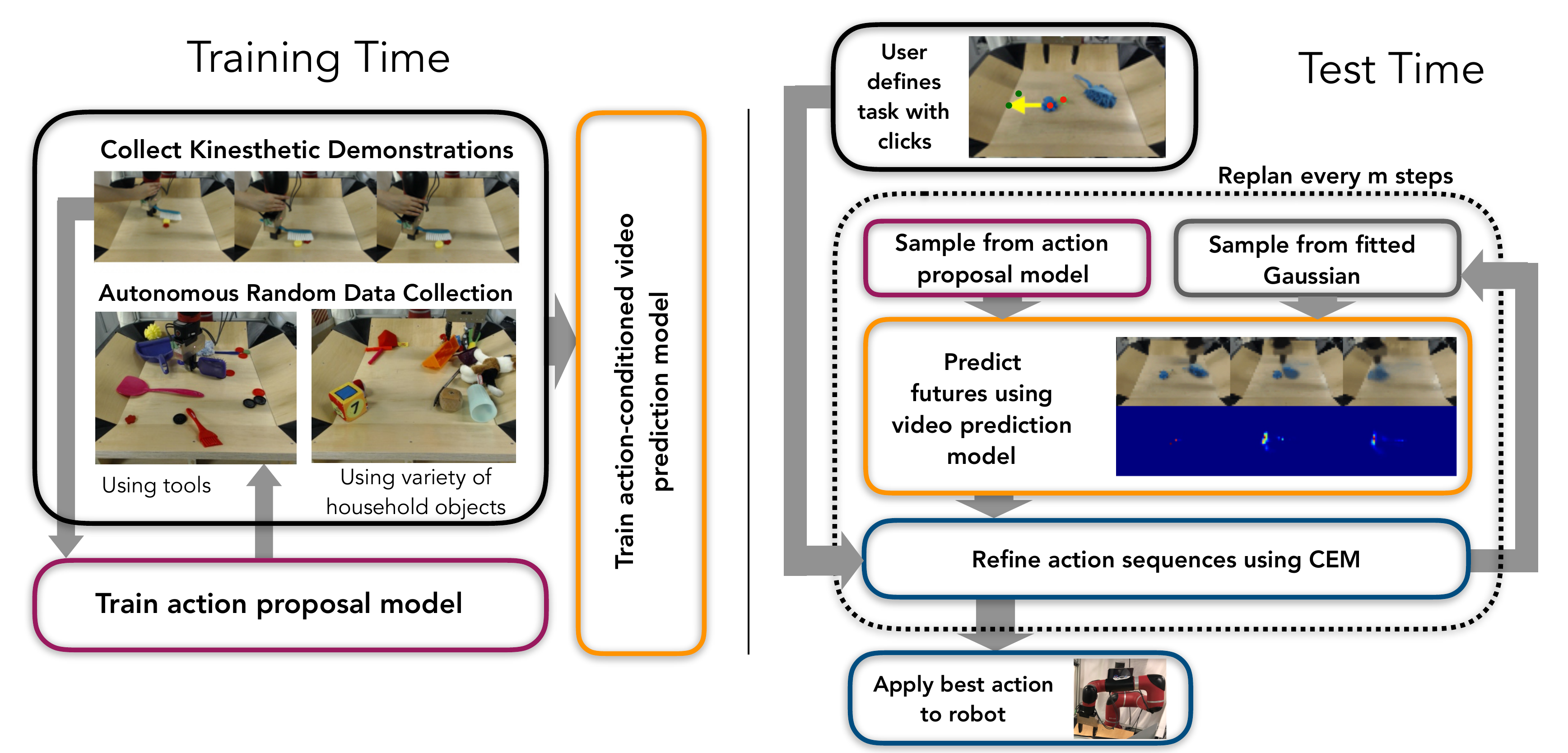}
    \vspace{-0.3cm}
    \caption{ Our guided visual foresight (GVF) approach, at training time (left) and test time (right). Our method incorporates demonstrations and autonomous data collection to learn a video prediction model and action proposal model that enable the robot to solve both a diverse range of goals that require tool use. We incorporate the action proposal model both for training data for the video prediction model and for improving the sampling-based planner at test time.
    The test time procedure is further detailed in Algorithm 1.}
    \vspace{-0.3cm}
    \label{fig:algorithm}
\end{figure*}

\begin{wrapfigure}{r}{0.45\linewidth} 
    \centering
    \includegraphics[width=0.9\linewidth]{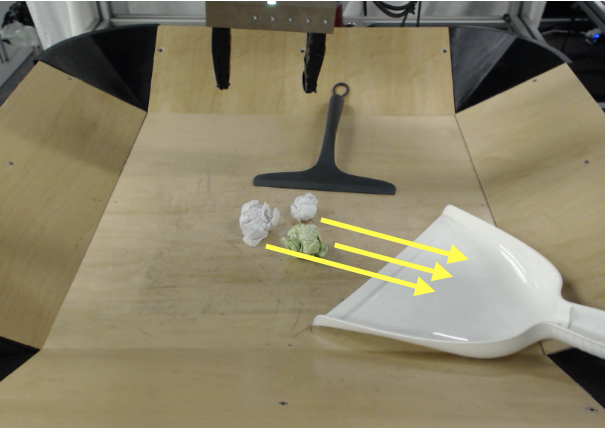}
    \vspace{-0.2cm}
    \caption{How can a robot use a tool that it has never seen before to accomplish the goal?
    }
    \vspace{-0.3cm}
    \label{fig:sweep_clutter}
\end{wrapfigure}
Our goal is to study how robots can use novel, previously-unseen objects as tools in order to perform tasks that cannot be completed without tools. See Figure~\ref{fig:sweep_clutter} for an example: the goal is for the robot to move the clutter onto the dustpan. Despite the robot having never previously seen the scraper, the clutter, or the dustpan in the scene, we would like the robot to figure out how it might use the scraper as a tool to efficiently clean the clutter.
We hypothesize that one way to accomplish this is for the robot to learn to predict the consequences of its actions and the outcomes of object-object interactions. That is, to learn \emph{generally} about how different objects interact with each other. Building realistic models of multi-object, non-prehensile interactions is challenging with analytic methods, as friction and contact dynamics become extraordinarily complex~\cite{fazeli2017empirical}, and inferring them directly from images is extremely difficult.
Further, such models are not extendable to previously unseen objects, without requiring detailed physical knowledge of the object.  In contrast, it is relatively easy for a robot to autonomously collect substantial amounts of data of object-object interactions.
Thus, if we are able to \emph{learn}
a model from such interaction data, we can then use such a model for planning to use tools.

However, even with a learned model, generalization remains a critical challenge. In order for a robot to be able to plan to use novel, previously-unseen objects as tools, the robot needs a representation that can effectively generalize to new characteristics of objects such as new sizes, shapes, and masses. How then should we represent the objects and the environment? One option is to represent objects and surfaces using 3D meshes or voxel grids. However, this puts significant stress on acquiring a robust perception system that generalizes to novel tools, and would likely require significant supervised or simulation data. An alternative option is to use raw sensor readings, such as image pixels, as the representation. While such a representation does not incorporate object-centric inductive biases, it does have a number of benefits. First, we can train models of sensory observations, i.e. video prediction models, from completely unlabeled interaction data requiring no manual annotation. Second, low-level sensory observations such as pixels include all information about the environment that the robot can currently perceive, and hence are general to a wide range of objects and situations, including nonrigid and deformable objects.
Motivated by these benefits, we will explore how we might enable improvisational tool use by autonomously collecting data of diverse object interactions, training predictive models of low-level sensory observations (i.e. action-conditioned video prediction~\cite{finn_nips}), and using these models to make plans to achieve goals involving tools. In the next section, we will describe how we can extend visual MPC~\cite{finn2017deep,ebert2018visual} to allow us to study such complex tasks.

%% file: method.tex
\section{Demonstration-Guided Visual Planning}

We aim to use demonstrations to better enable visual MPC, or visual foresight, to perform more complex, temporally-extended tasks. While demonstrations are typically used with single-task imitation learning, we hope to incorporate the demonstrations in a way that maintains the generality of visual MPC. That is, we want both breadth and depth: a method that can be used both for solving a \emph{variety} of tasks with unseen objects, and for solving a variety of \emph{complex} goals, such as picking up a tool and using it.
To this end, we will collect demonstrations that cover a broad range of tasks and goals using a range of tools. 
These demonstrations will be used in two ways:
for improving the video prediction model and for improving the sampling-based planning process. Specifically, we will use demonstrations, first, to enable the robot to collect data in parts of the state space that the robot would be unlikely to visit with random interaction, and second, to aid the sampling-based planner in finding solutions that are more difficult to find when searching from scratch.
We next give a short overview of our guided visual foresight approach (GVF), and then describe each of the components in more detail.

As shown in the left of \autoref{fig:algorithm} and described in~\autoref{sec:training_time}, the \emph{training time} procedure consists of three parts. We collect kinesthetic demonstrations of trajectories involving tool use. We use this data to train an \emph{action proposal model} to obtain a distribution over action sequences conditioned on the initial image based on actions taken by the demonstrator. This proposal model will be used both during training, for collecting data for improving the video prediction model, and at test time, to help warm-start the optimization over action sequences. In addition to the human demonstrations, the robot autonomously collects data by executing random actions. Finally, we train an action-conditioned video prediction model to predict future video sequences based on the initial image and the corresponding action sequences. 

As illustrated on the right in \autoref{fig:algorithm} and described in~\autoref{sec:test_time}, at \emph{test time}, we enable the robot to plan to use objects as tools as follows. First, a user can specify a goal by clicking on objects in the image and selecting where the corresponding pixels should move. For example, the user might specify that three pieces of trash need to be moved to a location within a dustpan.
Then, the current observation is passed to the action proposal model, which returns a sampling distribution that is used to sample a certain number of action sequences. These samples will usually correspond to different ways of interacting with objects in the scene --- in our case different ways of using objects in the scene as tools. We feed each of the sampled action sequences into the video prediction model to predict their outcome as a video. We then rank these predictions using a cost function determined by the human-specified goal, and refine the best samples further. Lastly, the robot recomputes action plans after several control cycles.

Because the demonstrations entail a wide variety of tools and goals, our experiments find that pure imitation learning struggles to capture the breadth of such a distribution. However, these demonstrations can be effectively used to guide the planning process towards tool-related behaviors, while the predictive model is used to fully construct and refine a sequence of actions for completing the task.

\subsection{Training Time}
\label{sec:training_time}
\subsubsection{Demonstration data collection}
\label{sec:demo_collection}
\begin{figure}[t]
    \centering
    \includegraphics[width=\linewidth]{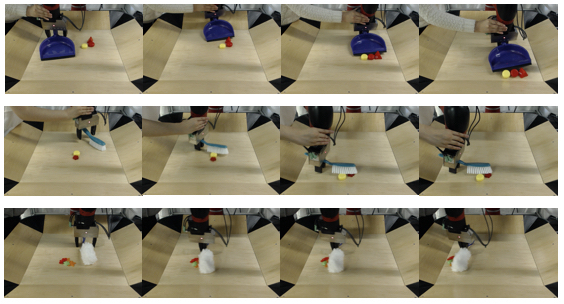}
    \vspace{-0.35cm}
    \caption{Examples of demonstrations collected via kinesthetic teaching. We provide demonstrations for a diverse range of tools, objects, and goals.}
    \vspace{-0.1cm}
    \label{fig:demo_data_coll}
\end{figure}

We collect demonstrations of tasks that require tools, typically involving grasping a tool and using the tool to maneuver other objects to a certain location. Because we specifically care not just about accomplishing a single task, but being able to perform a range of tasks with many different objects, we collect demonstrations for a variety of tasks that require a variety of tools (instead of many demonstrations for a single task).
In our prototype, we use kinesthetic teaching to collect demonstrations.
During the demonstrations we record images $I_t$, Cartesian end-effector positions $s_t$, as well as motor commands $a_t$ in the form of Cartesian end-effector displacements. We will denote the demonstration data as $\mathcal{D}_\text{demo} = \{ (I_1, s_1, a_1, I_2, s_2, a_2, ...)_j \}$
For examples of demonstrations, see \autoref{fig:demo_data_coll}. For each demonstration, we record a sequence of 24 to 30 time-steps. 
The tasks for these demonstrations are chosen such that success without tool use would be very low.

\subsubsection{Action proposal model training}
\label{sec:action_prop_training}

In order to collect additional data of interaction with tools and guide the planning process at test-time, we aim to acquire an approximate model of the tool-use behaviors seen in the demonstrations.
We fit an action proposal model $g_{\theta}$ to the demonstration data that outputs the distribution over a sequence of actions conditioned on the initial image $I_1$ and robot joint positions $s_1$. Note that outputting a distribution, rather than a deterministic output, will allow the model to capture a range of behaviors present in the diverse demonstrations. Because we would like to use this model for sampling-based planning, we do not condition the model on the final image of the demonstration.
The action proposal model is parameterized as an autoregressive recurrent neural network (RNN). It is trained with the following maximum likelihood objective:
\begin{equation}
    \max_{\theta} \!\!\! \sum_{I_1, s_1, a_{1:T} \in \mathcal{D_\text{demo}}} \!\!\! \log p_{\theta}(a_1, ..., a_T | I_1, s_1)
\end{equation}
Here, $I_1, s_1, a_{1:T}$ are the initial image observation, initial end-effector position, and action sequence in a kinesthetic demonstration, $p_{\theta}$ is mixture of Gaussians where the parameters are produced by $g_\theta$. 
We use a long short-term memory network (LSTM)~\cite{lstm} to model recurrence.
The model $g_\theta$ consists of three components: an initial state encoder $g_e$, an action encoder $g_a$, and an LSTM cell. The encoder $g_e$ encodes the initial image $I_1$ and state $s_1$ to provide the input to the LSTM at the first timestep, and the action encoder $g_a$ encodes the previous action to provide to the LSTM cell and future timesteps. The recurrent LSTM cell produces both the parameters of the Gaussian mixture and the next hidden state $h_{t+1}$. The full network is described by the following equations.
\begin{align}
    a_0 &= \mathbf{0}\\
    h_0, \varnothing &= \text{LSTM}(g_e(I_1, s_1), \mathbf{0}) ~~~ t=0 \\
    h_t, (\mu_t^{(i)}, \Sigma_t^{(i)}, w_t^{(i)}) &= \text{LSTM}(g_a(a_{t-1}), h_{t-1})  ~~\forall t>0\\
    p(a_{t}) &= \sum_{i \in \{1,..,N_{c}\}}  w_t^{(i)} \mathcal{N}(\hat{a}_{t}; \mu_t^{(i)}, \Sigma_t^{(i)}) \label{eq:action_prop_model_gmm}
\end{align}
The zero action is used to indicate the start of the sequence. 

\begin{figure}[t]
    \centering
    \includegraphics[width=\linewidth]{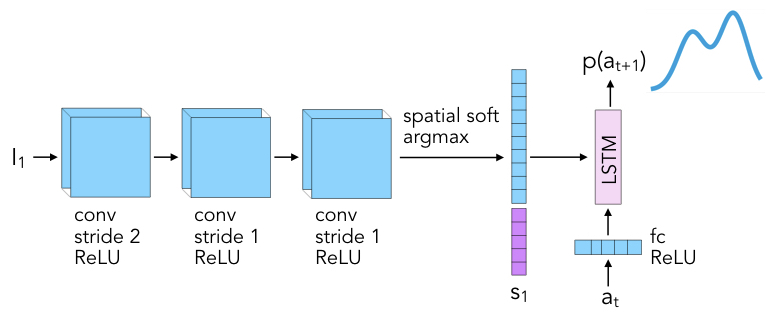}
    \vspace{-0.4cm}
    \caption{Architecture for the action proposal model. We use a recurrent autoregressive model to output the parameters of a Gaussian mixture model over the probbaility of an action at each timestep. Using recurrence and Gaussian mixtures enables the network to model diverse and multi-modal demonstration data based on the initial image.}
    \vspace{-0.1cm}
    \label{fig:action_prop_arch}
\end{figure}

We illustrate the neural network architecture for our action proposal model in Figure~\ref{fig:action_prop_arch}. To handle RGB image inputs, the network $g_e$ is composed of three convolutional layers with 32 3$\times$3 filters, each followed by batch normalization and a ReLU non-linearity. The first convolutional layer is initialized with pretrained weights from VGG-16. Then, spatial feature points are extracted from the last convolutional layer with a spatial soft-argmax operation~\cite{levine2016end}. We concatenate these features with the robot's initial end-effector position $s_1$ and pass them through a fully-connected layer with 50 units. As previously discussed, we use an RNN, in particular an LSTM net, that uses this embedding to initialize the hidden state. In the action encoder $g_a$, the actions are passed through a fully-connected layer, the result of which are the inputs to the LSTM. Finally, we use a mixture density network with 10 components to represent the output distribution~\cite{bishop1994mixture}.

\subsubsection{Autonomous data collection}
For training the predictive model, we need trajectory data $(I_1, s_1, a_1, I_2, s_2, a_2, ...)$ from the robot's interactions. The demonstration data, on its own, is quite limited in quantity. Hence, we also choose to have the robot collect data autonomously using, for example, random actions sampled from a Gaussian distribution. However, when training a video prediction model using this data, we observe that the model makes optimistic predictions, very often predicting that the robot will grasp a tool when approaching it, even when the grasp is not correctly positioned. To address this issue, the robot additionally collects data autonomously by sampling actions from the action proposal model. 

\begin{figure}[t]
    \centering
    \includegraphics[width=0.45\linewidth]{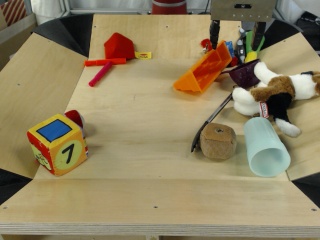}
    \includegraphics[width=0.45\linewidth]{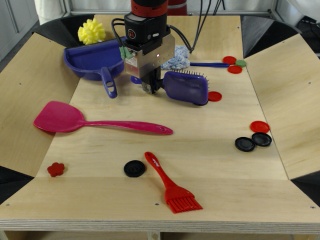}
    \vspace{-0.2cm}
    \caption{Autonomous data collection with a collection of house-hold objects (left) and a variety of of tools (right)}
    \vspace{-0.1cm}
    \label{fig:data_col}
\end{figure}

All in all, the video prediction model is trained on three sources of data: demonstrations, random interactions, and on-policy data.
When sampling the actions from the action proposal model a variety of tools are present in the robot workspace. When executing random actions, in some parts of the data we use tools, see \autoref{fig:data_col} (right), whereas in the other parts we use a collection of household items, shown in \autoref{fig:data_col} (left).
Further details about the composition of the dataset are given in \autoref{sec:experiments}.

To simplify picking up objects in general, including tools, we incorporate a simple ``grasp-reflex'' into the controller, where the gripper automatically closes when the height of the wrist above the table is lower than a small threshold (following~\citet{ebert2018visual}).
This reflex is inspired by the palmar reflex observed in infants~\cite{grasping_fetal}. With this primitive, when collecting data with random actions and rigid objects, about 20\% of trajectories included some sort of object grasp, significantly higher than without the reflex. Note, however, that this technique on its own is not sufficient for enabling tool use, as we find in the experiments.

\subsubsection{Predictive model training}
\label{seq:model_training}
Once we collect autonomous data, we use it to build a predictive model of future sensory inputs, i.e. images, conditioned on the initial image and the future actions taken.
We use a transformation-based video prediction architecture, first proposed by \citet{finn_nips}, and use the open-source architecture from~\citet{ebert2018visual}. The advantage of using transformation-based models over a model that directly generates pixels is two-fold: (1) prediction is easier, since the appearance of objects and the background scene can be reused from previous frames and (2) the transformations can be leveraged to obtain predictions about \emph{where} pixels will move, a property that is used in our planning cost function formulation, presented in \autoref{seq:planning_cost}. The model, which is implemented as a recurrent convolutional neural network, $f_{\gamma}$ parameterized by $\gamma$, has a hidden state $h_t$ and takes in a previous image and an action at each step of the rollout.  Future images $\hat{I}_{t+1}$ are generated by warping the previous generated image $\hat{I}_t$ or the previous true image $I_t$, when available, according to a 2-dimensional flow field $\hat{F}_{t+1 \leftarrow t}$. The forward pass of the dynamics model is summarized in the following two equations:
\begin{align}
[h_{t+1}, \hat{F}_{t+1 \leftarrow t}] 	&= f_{\gamma}(a_t, h_t, I_t) \\
\hat{I}_{t+1} 							&= \hat{F}_{t+1 \leftarrow t} \diamond  \hat{I}_t 
\label{eq:forward_model}
\end{align}
The model is trained with stochastic gradient descent using a $\ell_2$ image reconstruction loss. For more details on the architecture and training, see Appendix~\ref{app:videopred}.

\subsection{Test-Time Control}
\label{sec:test_time}

At test time, a user provides a goal to the robot, and the robot uses the learned action proposal and video prediction models to plan to achieve the goal. We describe this process in more detail next.

\subsubsection{Planning cost function}
\label{seq:planning_cost}
\begin{figure*}
\setlength{\unitlength}{0.5\columnwidth}
\vspace{-0.2cm}
\begin{picture}(1.99,1.1) \linethickness{0.5pt}
\put(0.0,0.0){\includegraphics[width=2.05\columnwidth]{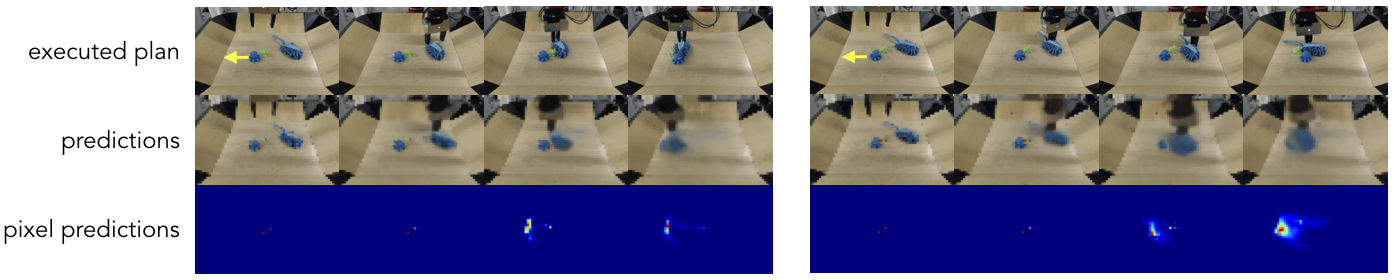}}
\end{picture}
\vspace{-0.3cm}
\caption{Examples of the lowest-cost predictions (2nd row) and executed actions (1st row), for the task indicated in the top left image. The left example shows a model trained on on-policy data, while the right example shows the best action sequence found with a model that was \emph{not} trained on demonstration data nor on data from the action proposal model. Note that the robot fails to grasp the object in the second example, while the model predicted that the grasp would be successful. Each example also shows the probability distribution of the designated pixel over time (3rd row). }
\vspace{-0.2cm}
\label{fig:tool_use_plans}
\end{figure*}

A user can provide a goal by clicking on a pixel corresponding to an object and a corresponding goal position for that pixel. A \emph{pixel distance cost function} evaluates how far the designated pixel is from the goal pixels. Given a distribution over pixel positions $P_0$, our model predicts distributions over its positions $P_t$ at time $t \in \{ 1, \dots, T \}$ as follows: 
To predict the future positions of the designated pixel $d$, the same transformations used to transform the images are applied to the distribution over designated pixel locations. The warping transformation $\hat{F}_{t+1 \leftarrow t}$ can be interpreted as a stochastic transition operator allowing us to make probabilistic predictions about future locations of individual pixels:
\begin{equation}
\hat{P}_{t+1} = \hat{F}_{t+1 \leftarrow t} \diamond  \hat{P}_t
\label{eqn:prob_forward}
\end{equation}
Here, $P_t$ is a distribution over image locations which has the same spatial dimension as the image (an example is shown in \autoref{fig:tool_use_plans} in the third row). For simplicity in notation, we will use a single designated pixel moving forward, but using multiple designated pixels is a straightforward extension. At the first time step, the distribution $\hat{P}_0$ is defined to be 1 at the position of the user-selected designated pixel and zero elsewhere.
One way of defining the cost per time-step $c_t$ is by using the expected Euclidean distance to the goal point $d_g$, which is straight-forward to calculate from $P_t$ and $g$, as follows:
 \begin{align}
c = \sum_{t = 1, \dots, T} c_t =  \sum_{t = 1, \dots, T} \mathbb{E}_{\hat{d}_{t} \sim P_{t}} \left[\|\hat{d}_{t} - d_{g}\|_2\right] 
 \label{eq:cost}
 \end{align}
The per time-step costs $c_t$ are summed together giving the overall planing objective $c$. For tasks with multiple designated pixels $d^{(i)}$, the costs are also summed together.

\subsubsection{Planning with demonstration guidance}
\label{sec:planning_with_demo}

\begin{algorithm}[t!]
\caption{Guided visual foresight (test time)}
\label{alg:1}
\begin{algorithmic}[1]
\State \textbf{Input:} Predictive model $f_{\gamma}$
\State \textbf{Input}: Planning cost $c$ derived from user-specified pixel goals
\For{$i~=~0...n_{iter}-1$}
\If{$i==0$}
\State \begin{varwidth}[t]{\linewidth}
	Sample $M$ action sequences $\{a^{(m)}_{1:H}\}$ from \par
	action proposal distribution by rolling out $g_{\theta}$
\end{varwidth}
\Else
\State \begin{varwidth}[t]{\linewidth}
	Sample $M$ action sequences ${a^{(m)}_{1:H}}$ from \par 
	$\mathcal N(\mu^{(i)}, \Sigma^{(i)})$
\end{varwidth}
\EndIf
\State  \begin{varwidth}[t]{\linewidth}
	Use $f_{\gamma}$ to predict future  image sequences $\hat{I}_{1:H}^{(m)}$\\ and probability distributions $\hat{P}_{1:H}^{(m)}$
\end{varwidth}
\State Rank action sequences using cost function $c$
\State  \begin{varwidth}[t]{\linewidth}
	Fit a Gaussian to the $k$  action samples\\ with lowest cost,
	yielding $\mu^{(i+1)}, \Sigma^{(i+1)}$
\end{varwidth}
\EndFor
\State Execute lowest-cost action sequence on the robot
\end{algorithmic}
\end{algorithm}

Planning with GVF at test time is illustrated in \autoref{fig:algorithm} (right) and Algorithm \autoref{alg:1}. The user first specifies the task by clicking on the pixels that shall be moved and the corresponding goal-pixels. The planner searches for actions using the cross entropy method (CEM) \cite{kroese2013cross}, a common iterative sampling-based optimization procedure. To allow the optimizer to find more complicated, temporally extended action sequences, such as picking up a tool and using it in a goal-directed manner, we sample actions from the stochastic action proposal model $g_{\theta}$ in the first iteration of CEM, as listed in line 5 of Algorithm \autoref{alg:1}. After rolling out the video prediction model $f_{\gamma}$ using \autoref{eq:forward_model} we obtain $M$ different predicted probability distributions $\hat{P}^{m}_{1:H}$, which are ranked using the cost function $c$. We then fit a Gaussian distribution to the best $k$ action samples (see line 10). In later CEM iterations, actions are sampled from the fitted Gaussians (line 7).
In practice, we choose the number of samples $M$ to be 100, the horizon $H$ to be 21, and the number of CEM iterations $n_{iter}$ to be 3. 

\begin{figure}[t]
    \centering
    \includegraphics[width=0.75\linewidth]{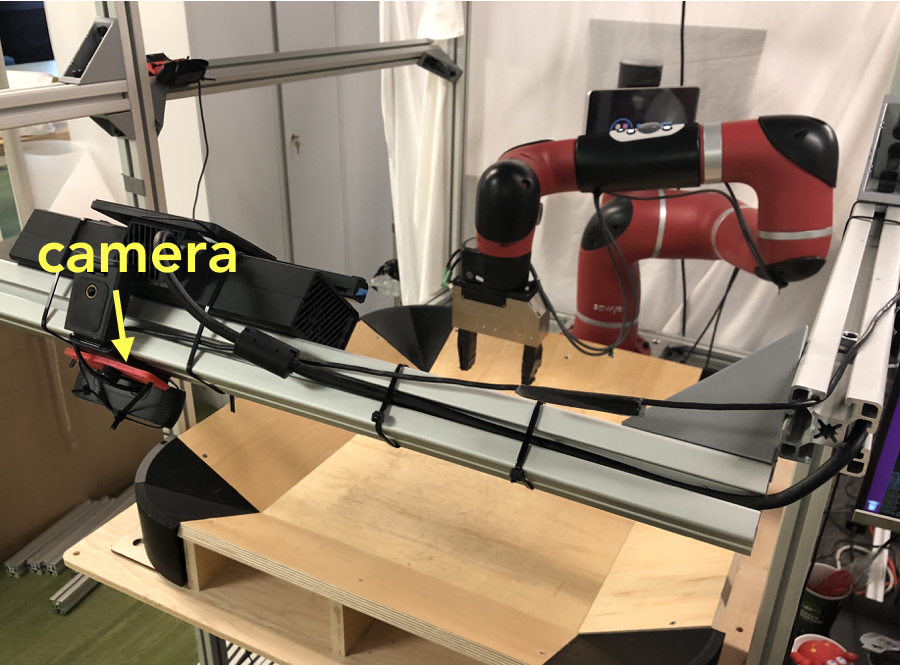}
    \vspace{-0.25cm}
    \caption{The physical robot set-up: we use a Sawyer robot with Cartesian space impedance control to ensure soft interaction with objects. RGB images are taken from a conventional webcam, as indicated in the image.}
    \vspace{-0.3cm}
    \label{fig:setup}
\end{figure}

%% file: experiments.tex
\section{Experiments}
\label{sec:experiments}
In our experiments, we aim to answer the following questions:
\textbf{(1)} can our approach effectively solve tasks that require tool use? \textbf{(2)} can our method improvise, by figuring out how to use a new object that was not seen during training as a tool? \textbf{(3)} how important is the action proposal model? \textbf{(4)} can our method dynamically decide to use or not use tools depending on the demands of the task?
To answer these questions, we conduct experiments on a Sawyer robotic arm, with an experimental set-up shown in Figure~\ref{fig:setup}. Video results are in the supplemental materials and the supplementary webpage\footnote{For videos, see \url{https://sites.google.com/view/gvf-tool}}.

\begin{figure}[t]
    \centering
    \includegraphics[width=0.75\linewidth]{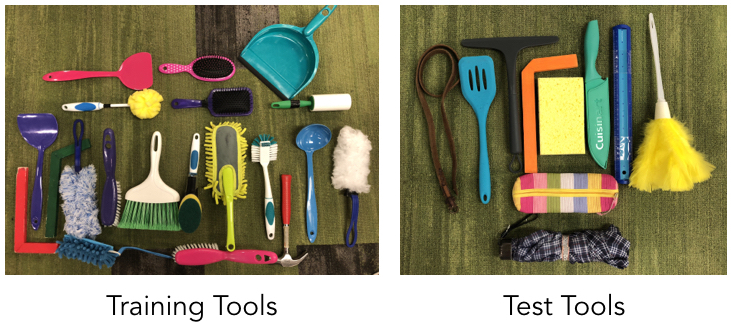}
    \vspace{-0.25cm}
    \caption{Left: tools used during training. Right: test tools used in our quantitative evaluation, some of which are not conventional tools.}
    \vspace{-0.1cm}
    \label{fig:tools}
\end{figure}

\begin{figure}[t!]
    \centering
    \includegraphics[width=0.51\linewidth]{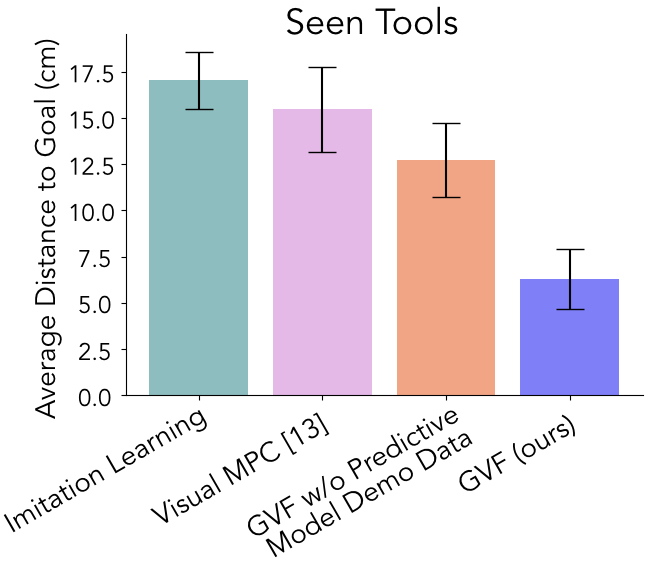}
    \hspace{-0.45cm}
    \includegraphics[width=0.51\linewidth]{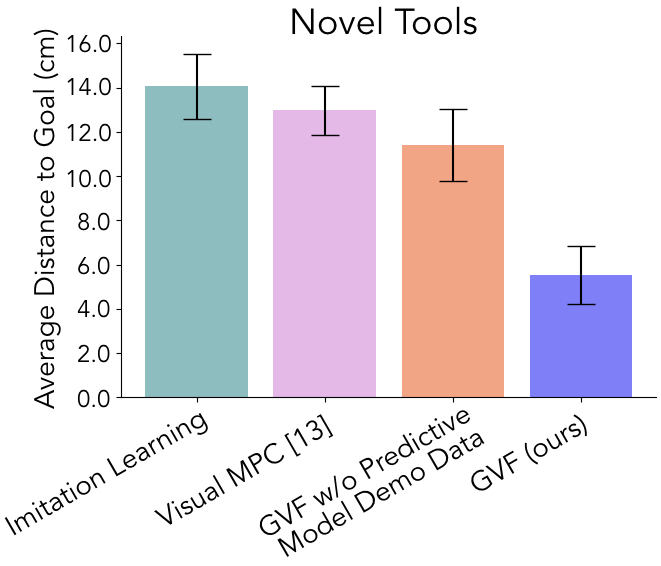}
    \vspace{-0.6cm}
    \caption{Quantitative results: our approach, which uses elements of imitation learning and visual MPC, significantly outperforms either approach individually. In particular, we compare to direct imitation learning on the diverse demonstration data, to visual MPC without the learned action proposal model, and to our method with a video prediction model trained only on autonomously collected data.}
    \vspace{-0.2cm}
    \label{fig:sawyer_quant_results}
\end{figure}

\begin{figure}[t]
    \centering
    \includegraphics[width=\linewidth]{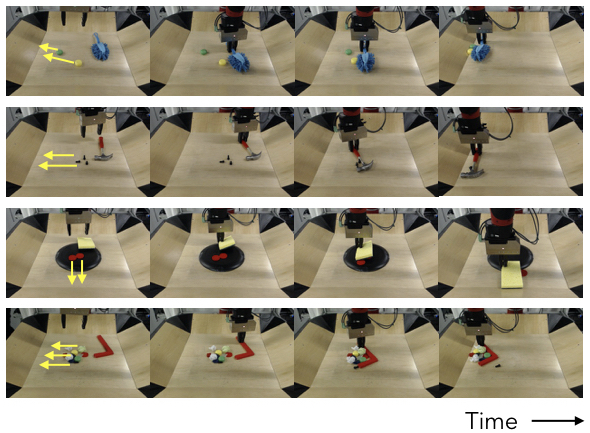}
    \vspace{-0.65cm}
    \caption{Qualitative results illustrating our approach (GVF): the robot executing the lowest-cost plan according to the pixel-distance cost function. The arrows in the left column of images indicate where the robot is to maneuver the objects.}
    \vspace{-0.3cm}
    \label{fig:sawyer_qual_results}
\end{figure}

\subsection{Experimental Set-Up and Comparisons}
To train the action proposal model discussed in Section~\ref{sec:action_prop_training}, we collected kinesthetic demonstrations on a Sawyer robotic arm with twenty different tools. Figure~\ref{fig:tools} (left) illustrates these twenty tools. Here, we focus on sweeping, scraping, and wiping tasks where the goal is to move multiple objects which would be infeasible to complete without the use of a tool. Thus, in each demonstration, we randomly place a tool and pile of objects in front of the robot and kinesthetically demonstrate how to grasp the tool and sweep the smaller objects (see \autoref{fig:demo_data_coll}). The demonstrations are recorded at 5 Hz and range from 24 to 30 time steps. The action proposal model is then trained on subsequences of 10 steps, conditioned on the image observation and robot state of the first step of the subsequence. 

As described in Section~\ref{seq:model_training}, to train the video prediction model, we collect additional interaction data by taking random actions and by rolling out samples from our action proposal model. Our final dataset is composed of: 16,000 random trajectories from the open-source dataset in \cite{ebert2018robustness}, 5,052 random trajectories with tools, 1,754 samples from the action proposal model, and 1,200 demonstrations. 

To study the importance of visual planning and the importance of demonstrations in both video prediction model training and planning, we compare our method to the following approaches:
\begin{itemize}[leftmargin=*]
    \item \textbf{Imitation Learning}: Sampling from an action proposal model that is conditioned on the initial and goal image observations, representative of standard imitation learning~\cite{alvinn,atkeson1997robot,codevilla2018end,pathak2018zero}. This comparison evaluates the importance of physical prediction and planning.
    \item \textbf{Visual MPC}: 
    Our method with CEM samples from a Gaussian distribution. This comparison is representative of visual MPC~\cite{finn2017deep,ebert2018visual}. The video prediction model is still trained with demonstrations and samples from the action proposal model, so it is actually stronger than the method of~\citet{ebert2018visual}. This comparison evaluates the importance of demonstrations in guiding the planning process.
    \item \textbf{GVF w/o Predictive Model Demo Data}: Our method with a video prediction model trained only on randomly collected data, omitting demonstration data and data from the action proposal model when training the video prediction model. Test-time planning still uses the action proposal model. This comparison evaluates the role of demonstrations in improving the predictions of multi-object interactions.
\end{itemize}
The imitation learning method uses a convolutional neural network with the same architecture as for the action proposal model, described in Section~\ref{sec:action_prop_training} to map both the initial and goal image observations. The feature points from both image inputs, along with the robot's initial end-effector position, are concatenated and passed through a fully-connected layer of 50 units. The output is then used as the initial state of the LSTM.

\begin{figure}[t]
    \centering
    \includegraphics[width=\linewidth]{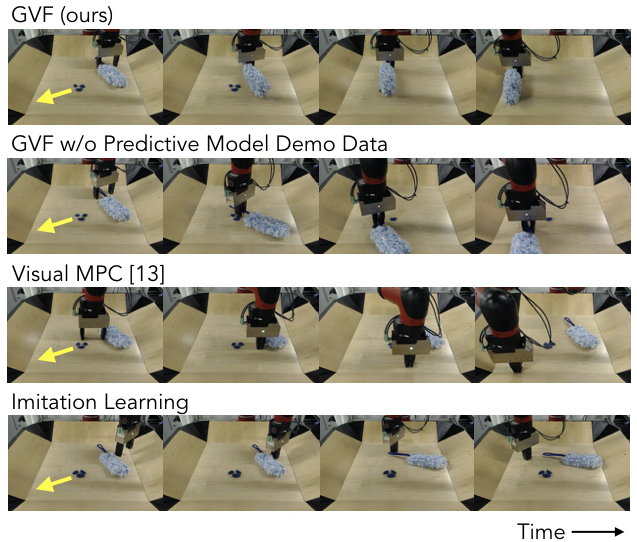}
    \vspace{-0.5cm}
    \caption{Qualitative comparison of our method to prior methods and ablations. Without the action proposal model (visual MPC), the method generally cannot find the actions that grasp the tool, while without the video prediction model, imitation learning generally fails to use the tool in a meaningful way.}
    \vspace{-0.2cm}
    \label{fig:qual_comparison}
\end{figure}

\subsection{Experimental Results}

We quantitatively evaluate each method on 10 tasks with tools seen during training and 10 tasks with previously unseen tools, with results summarized in Figure~\ref{fig:sawyer_quant_results} and detailed in Appendix~\ref{app:detailed_results}.  Each task requires picking up the tool and sweeping, scraping, or wiping objects to the position corresponding to the specified goal pixels. Note that the set of tasks with seen tools differs from the set of tasks with novel tasks, thus the two sets of results are not directly comparable. In regard to question \textbf{(1)}, these results show that our method can successfully use tools when they are available, reaching less than half the position error to the goal compared to the prior methods. The qualitative results and supplementary video show that the robot is generally successful at the tool use tasks. In regard to question \textbf{(2)}, the relatively similar performance with novel tools indicates that the robot can generalize effectively, utilizing new objects as tools when needed. The qualitative results illustrate examples of novel objects that the robot can use as tools.

We further evaluate questions \textbf{(1)} and \textbf{(2)} through qualitative results and comparisons.  Our primary qualitative results are in Figures~\ref{fig:teaser} and~\ref{fig:sawyer_qual_results}, including both seen and novel tools. As illustrated in Figure~\ref{fig:sawyer_qual_results} (third row), we find that, even though the robot has never seen a sponge before, the robot is able to use the sponge to wipe a plate.  Further, as shown in Figure~\ref{fig:teaser} (second row), the robot has never seen or interacted with a knife before, it can figure out how to use the knife to push two red pieces of rubbish to the edge of the cutting board. Finally, when no conventional tool is available, the robot is able to improvise when tasked with moving pieces of trash, by grasping a water bottle and using it to sweep the trash to the side (see Figure~\ref{fig:teaser}, third row).

In regard to question \textbf{(3)}, incorporating demonstration data for both the sampling distribution and training the video prediction model on average leads to more successful behavior. Moreover, compared to learning a predictive model with demonstration data, the action proposal model results in better performance on our set of evaluation tasks. We analyze the failure modes of the prior methods and ablations in Figure~\ref{fig:qual_comparison}.

\begin{figure}[t]
    \centering
    \includegraphics[width=\linewidth]{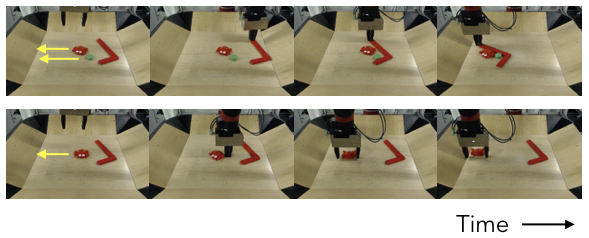}
    \vspace{-0.7cm}
    \caption{Our approach solves a task using a tool where it needs to manipulate two objects simultaneously (top), and chooses to not use a tool when the task involves only one object, allowing the robot to complete the task more efficiently (bottom).}
    \vspace{-0.2cm}
    \label{fig:no_tool_use}
\end{figure}

Lastly, in regard to question \textbf{(4)}, we aim to test whether our method still retains the full \emph{generality} of visual MPC by evaluating whether it can plan to solve a task \emph{without} using a tool, e.g. when tool-use does not have an advantage, or even a disadvantage. Therefore we set up the following experiment: as shown in \autoref{fig:no_tool_use}, we have two almost identical task settings, where the only difference is that in the first task two objects need to be pushed and in the second task only one object needs to be moved. In the latter case using a tool does not have any advantage, therefore we expect the planner to find a plan that does not use the tool. To allow the planner to explore both tool-use and non-tool use options, in the first CEM-iteration we use a combination of 50 samples from the proposal model and 50 samples from a unit Gaussian. As shown in \autoref{fig:no_tool_use}, GVF is indeed able to find a non-tool use trajectory for the single-object pushing task.

%% file: appendix.tex
\clearpage
\newpage
\section{Appendix}

\vspace{-0.1cm}
\subsection{Video Prediction Model Implementation details}
\label{app:videopred}
\vspace{-0.1cm}

\begin{figure}[b]
\vspace{-0.2cm}
	\centering
	\includegraphics[width=0.9\columnwidth]{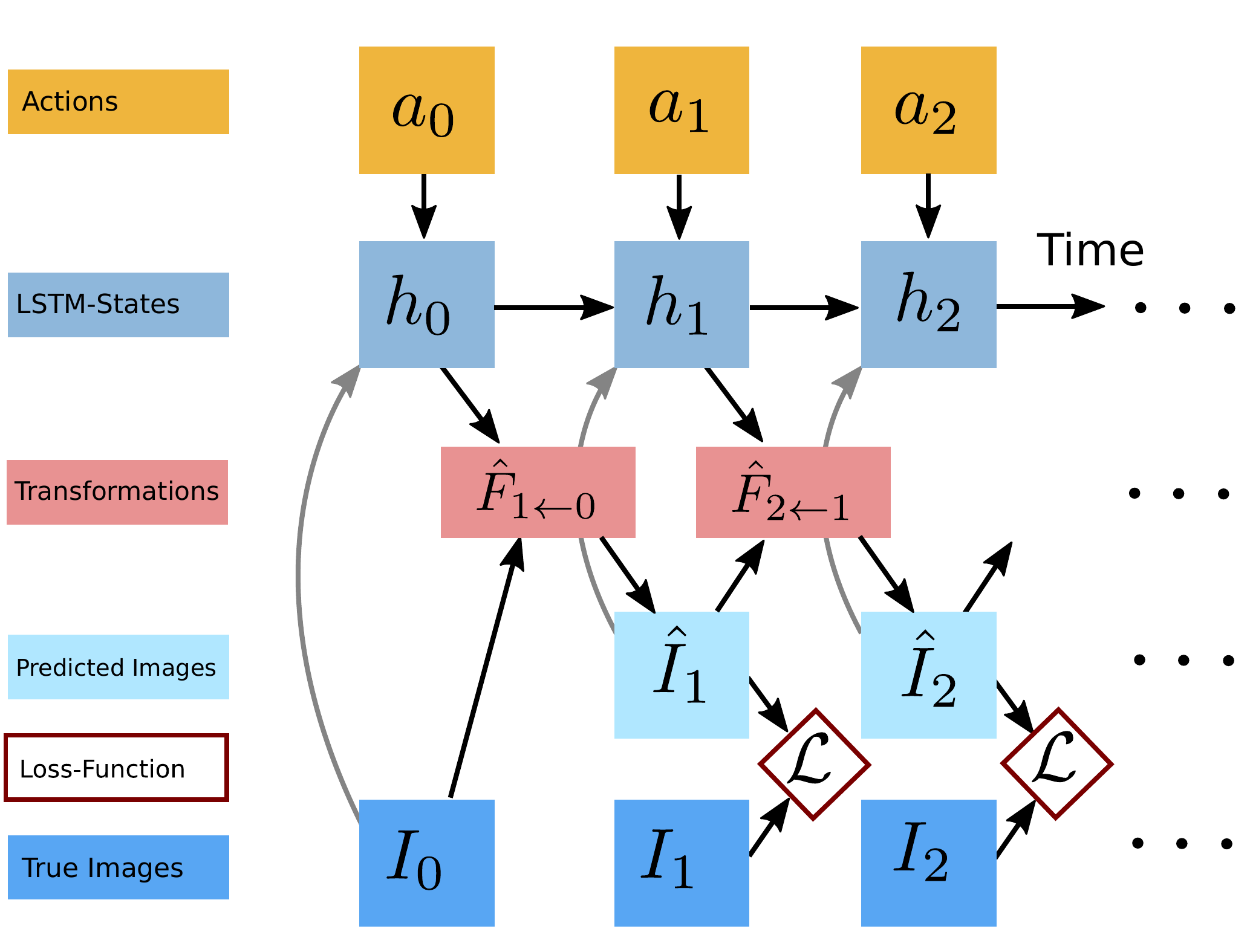}
	\vspace{-0.2cm}
	\caption{\small{Computation graph of the video prediction model. Time goes from left to right, $a_t$ are the actions, $h_t$ are the recurrent hidden states, $\hat{F}_{t+1 \leftarrow t}$ is a 2D-warping field, $I_t$ are real images, and $\hat{I}_t$ are predicted images, $\mathcal{L}$ is a pairwise training-loss. In this illustration $I_0$ is a context frame. Used with permission from \citet{ebert2018visual}. }}   
	\label{fig:prediction_model}
\end{figure}

We deterministic variant of the video prediction model described in \cite{savp}. The video prediction model is realized as a transformation-based model which generates future images by \emph{transforming} past images. The core of the model is made up from a recurrent neural network, \autoref{fig:prediction_model} gives an overview of a roll-out through time. In practice, the first two images passed into the model are ground truth images, called \emph{context frames}. At every time-step an action $a_t$ is passed into the model along with the hidden state $h_t$, producing a new state $h_{t+1}$ and a flow field $\hat{F}_{t+1 \leftarrow t}$ which is used to to transform the image via bi-linear sampling.

\begin{figure}[t]
    \centering
    \includegraphics[width=\columnwidth]{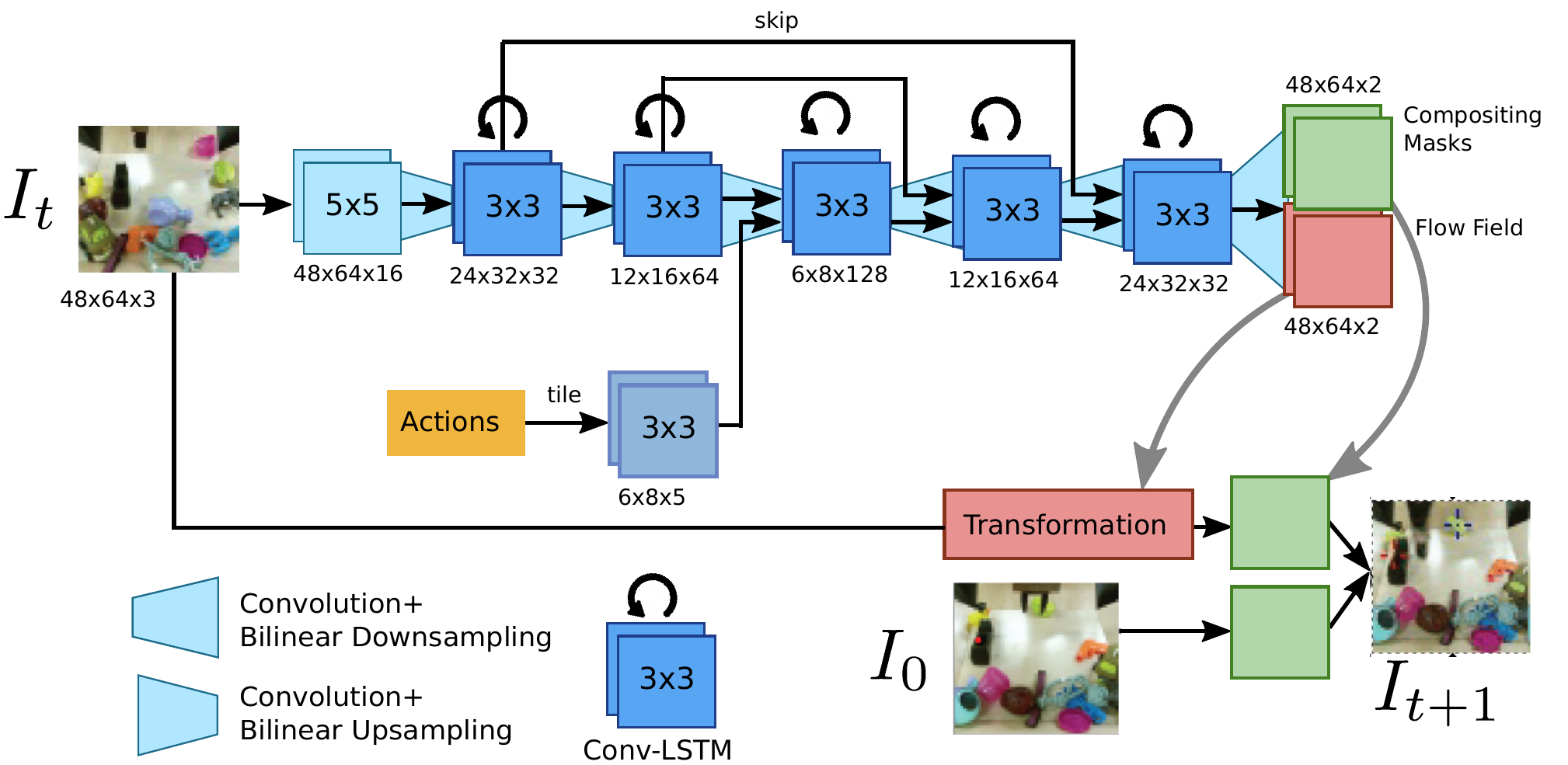}
    \vspace{-0.2cm}
    \caption{\small{Forward pass through the recurrent SNA model. The image from the first time step $I_0$ is concatenated with the transformed images $\hat{F}_{t+1 \leftarrow t} \diamond  \hat{I}_t $ multiplying each channel with a separate mask to produce the predicted frame for step $t+1$. Used with permission from \citet{ebert2018visual}. }}       \label{fig:occlusion_model}
\end{figure}

\autoref{fig:occlusion_model} shows the forward pass of a single time-step. The network consists of multiple layers of convolutional LSTMs \cite{convlstm}, a spatial, convolutional version of standard LSTMS, which are more efficient computationally and provide a regularizing inductive bias. While the transformations in theory would be sufficient to predict most parts of a video, it was found that allowing the model to selectively copy parts of the image from the \emph{first frame} of the sequence helps overcoming problems that occur with \emph{occluding objects}, i.e. objects in the fore-ground, would erase other parts of the image when they are moving \cite{ebert2017self}. Copying parts from the first image is achieved by predicting compositing masks (shown in green), a set of features maps with the same size of the image passed through a channel-wise softmax so that they add up to one along the channel-dimension.

The prediction loss is implemented as a standard $l1$-error. To regularize the RNN scheduled sampling \cite{bengio2015scheduled} is used, which provides a training curriculum for more stable RNN training.
The model is trained for 300k steps with standard backpropagation through time (BPTT), for the optimizer we use Adam.
For more details concerning the video prediction model implementation we refer the reader to \cite{ebert2018visual} and \cite{savp}.

\subsection{Detailed Quantitative Task Results}
\label{app:detailed_results}

In Tables~\ref{tbl:seen_tools_results} and~\ref{tbl:novel_tools_results}, we show the complete results for all 20 evaluated tasks.

\begin{figure}[h]
\small
\centering
\begin{tabular}{|c|c|c|c|c|}
\hline
Task & \begin{tabular}[c]{@{}c@{}}GVF\\ (ours)\end{tabular} & \begin{tabular}[c]{@{}c@{}}GVF w/o \\ Predictive Model\\ Demo Data\end{tabular} & \begin{tabular}[c]{@{}c@{}}Visual \\ MPC~\cite{ebert2018visual}\end{tabular} & \begin{tabular}[c]{@{}c@{}}Imitation\\ Learning\end{tabular} \\ \hline
1 & 10.0 & 20.0 & 20.0 & 18.5 \\ \hline
2 & 8.5 & 7.0 & 8.5 & 14.5 \\ \hline
3 & 0.5 & 13.0 & 15.7 & 15.7 \\ \hline
4 & 9.5 & 5.5 & 19.3 & 18.5 \\ \hline
5 & 1.0 & 16.7 & 24.3 & 24.3 \\ \hline
6 & 0.3 & 8.0 & 7.5 & 14.5 \\ \hline
7 & 5.0 & 18.3 & 22.0 & 18.3 \\ \hline
8 & 13.5 & 18.7 & 18.7 & 20.0 \\ \hline
9 & 13.7 & 18.7 & 18.7 & 21.0 \\ \hline
10 & 1.7 & 1.7 & 0.5 & 5.0 \\ \hline
\end{tabular}
\label{tbl:seen_tools_results}
\caption{Average distance to goal (in centimeters) for each evaluation task with previously seen tools.}
\end{figure}

\begin{figure}[h]
\small
\centering
\begin{tabular}{|c|c|c|c|c|}
\hline
Task & \begin{tabular}[c]{@{}c@{}}GVF\\ (ours)\end{tabular} & \begin{tabular}[c]{@{}c@{}}GVF w/o \\ Predictive Model\\ Demo Data\end{tabular} & \begin{tabular}[c]{@{}c@{}}Visual \\ MPC~\cite{ebert2018visual}\end{tabular} & \begin{tabular}[c]{@{}c@{}}Imitation\\ Learning\end{tabular} \\ \hline
1 & 3.0 & 12.8 & 11.5 & 14.0 \\ \hline
2 & 8.3 & 10.0 & 15.5 & 10.0 \\ \hline
3 & 2.0 & 0.0 & 9.5 & 9.0 \\ \hline
4 & 1.5 & 19.5 & 18.0 & 21.5 \\ \hline
5 & 4.5 & 11.0 & 12.0 & 11.5 \\ \hline
6 & 12.5 & 19.3 & 19.3 & 18.8 \\ \hline
7 & 3.0 & 9.5 & 8.0 & 13.0 \\ \hline
8 & 0.3 & 11.0 & 11.5 & 11.0 \\ \hline
9 & 10.5 & 11.0 & 14.5 & 22.0 \\ \hline
10 & 10.0 & 10.0 & 10.0 & 10.0 \\ \hline
\end{tabular}
\label{tbl:novel_tools_results}
\caption{Average distance to goal (in centimeters) for each evaluation task with novel tools.}
\end{figure}